\documentclass[10pt,twocolumn,letterpaper]{article}

\usepackage{cvpr}
\usepackage{times}
\usepackage{epsfig}
\usepackage{graphicx}
\usepackage{amsmath}
\usepackage{amssymb}
\usepackage{float}
\usepackage{multirow}
\usepackage[misc]{ifsym}

\usepackage[breaklinks=true,bookmarks=false]{hyperref}

\cvprfinalcopy 

\def\cvprPaperID{****} 

\begin{document}

\title{The Direction-Aware, Learnable, Additive Kernels and the Adversarial Network for Deep Floor Plan Recognition}

\author{Yuli Zhang\textsuperscript{1}\\
Technique Center Ihome Corporation\\
Shenzhen\\
ZHYL765@163.com\\
\and
Yeyang He\textsuperscript{2}\\
Technique Center Ihome Corporation \\
Shenzhen\\
861110348@qq.com\\
\and
Shaowen Zhu\textsuperscript{3}\\
Technique Center Ihome Corporation \\
Nanjing\\
shaowen5011@gmail.com\\
\and
\Letter Xinhan Di\textsuperscript{1}\\
Technique Center Ihome Corporation \\
Nanjing\\
deepearthgo@gmail.com\\
}
\maketitle
\begin{abstract}
This paper presents a new approach for the recognition of elements in floor plan layouts. Besides of elements with common shapes, we aim to recognize elements with irregular shapes such as circular rooms and inclined walls. Furthermore, the reduction of noise in the semantic segmentation of the floor plan is on demand. To this end, we propose direction-aware, learnable, additive kernels in the application of both the context module and common convolutional blocks. We apply them for high performance of elements with both common and irregular shapes. Besides, an adversarial network with two discriminators is proposed to further improve the accuracy of the elements and to reduce the noise of the semantic segmentation. Experimental results demonstrate the superiority and effectiveness of the proposed network over the state-of-the-art methods.   

\end{abstract}

\begin{figure*}
	\centering
	\includegraphics[width=12cm]{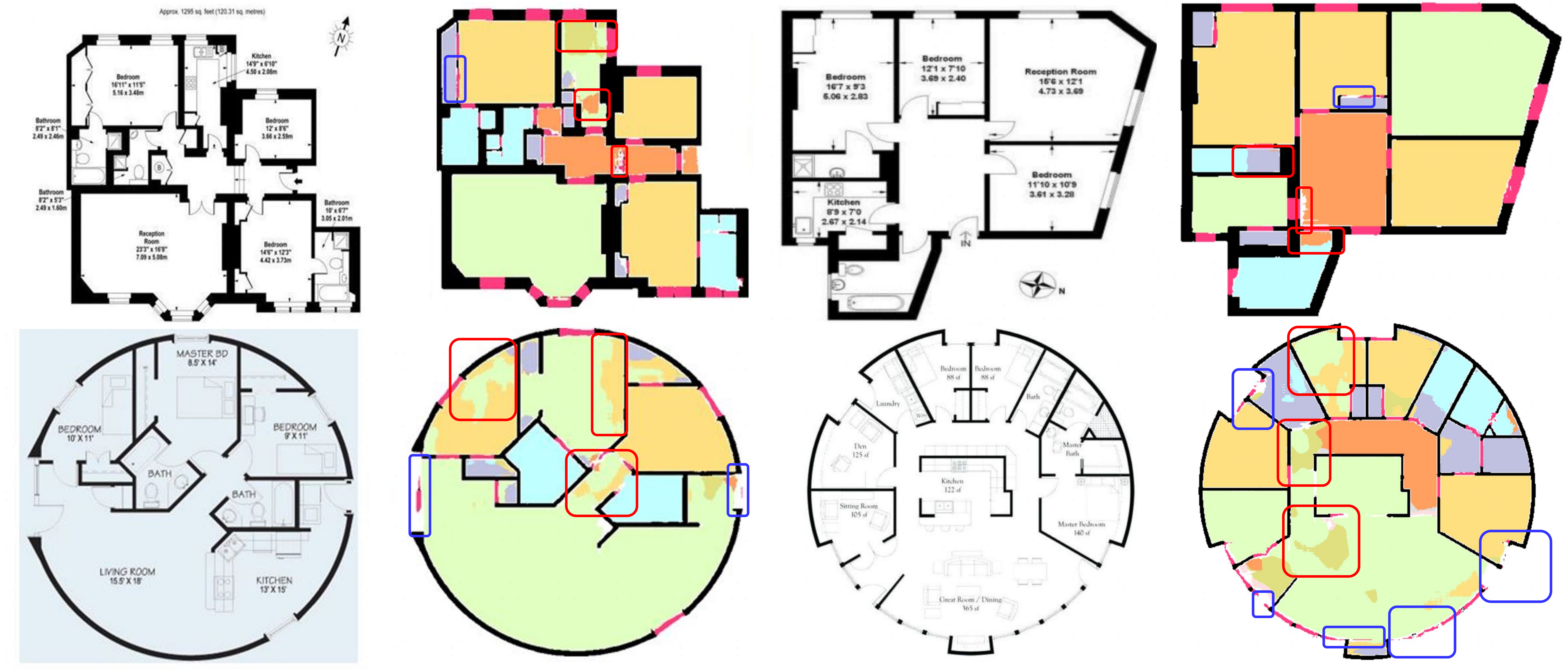}
	\caption{Noisy results of the floor plan with irregular shapes. The red bbox represents noise in the room. The blue bbox represents the noise in the boundary.}
	\label{fig1}
\end{figure*}

\section{Introduction}
The learning for the recognition of the floor plan elements is a principal task of the automatic interior decoration. The recognition of the 2D floor plan elements provides significant information for the automatic furniture layouts in the 3D world \cite{zeng2019deep}. This task is both relative with the general segmentation and the relation among the floor plan elements. In detail, the segmentation of the elements such as walls, doors, windows is the basic step. The understanding of the relation among these elements is also required for obtaining satisfactory result. However, the automatic processing of the floor plans and recognition of the layout semantics is a very challenging problem in the image understanding and processing. Besides, in practice, the performance of the recognition is far beyond satisfactory as there are a variety of irregular rooms in the real world.

\begin{figure}
	\centering
	\includegraphics[width=5cm]{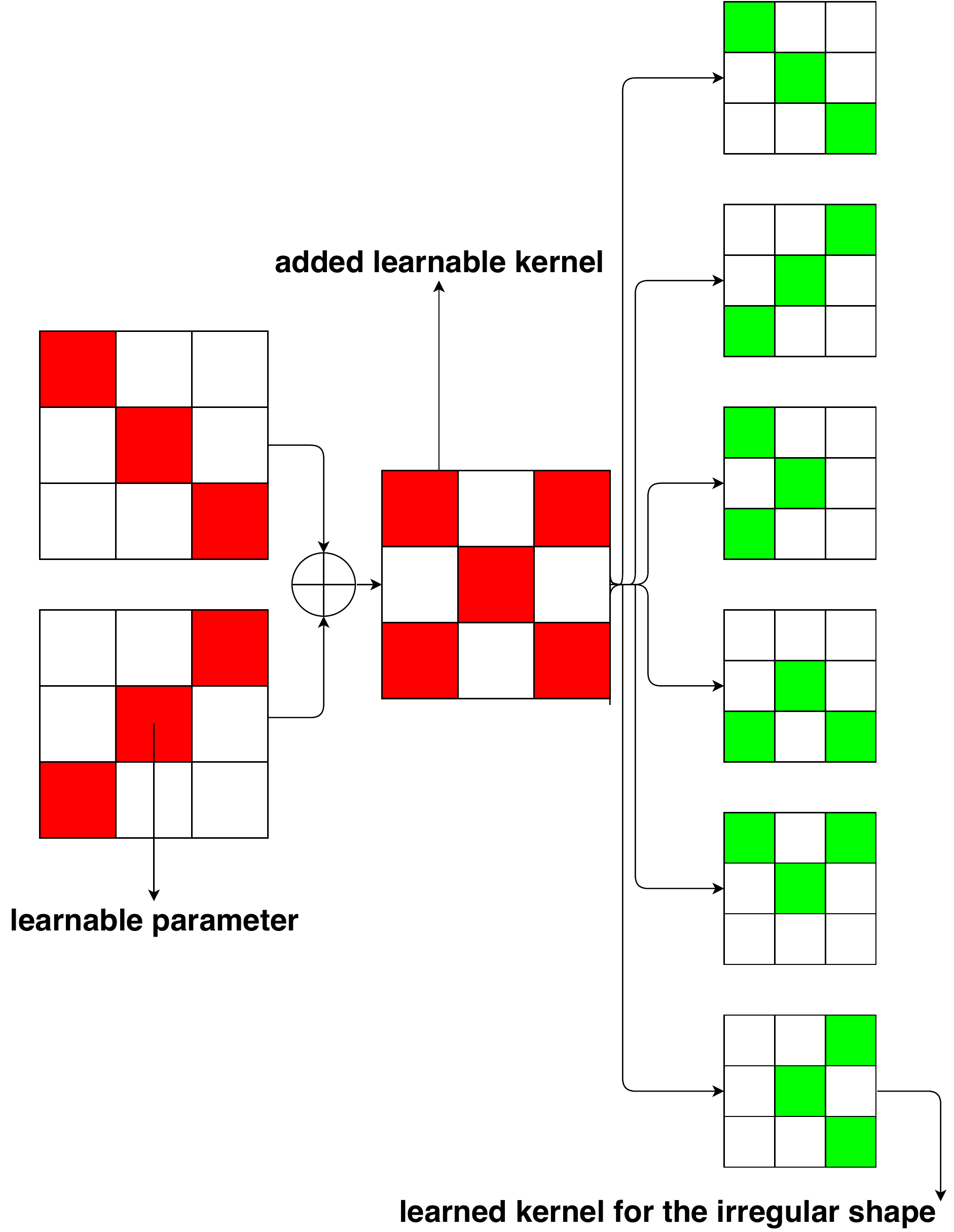}
	\caption{Direction-aware, learn-able, additive kernels. The parameters in the kernels is learn-able, the two kernels are added as one learn-able kernel. The kernel adapts to direction-aware kernels for different elements with the both common and irregular shapes. }
	\label{fig2}
\end{figure}

Traditional methods use the low-level image processing \cite{mace2010system,ahmed2011improved,gimenez2016automatic}, heuristics and semantics are exploited to locate the graphical notations in the floor plans. However, these simple hand-crafted features and extraction of these features are insufficient, as the generality of these features is limited to handle with diverse conditions in the real world.

Therefore, deep learning approaches are proposed to explore this problem. A convolutuional neural network (CNN) is developed to recognize junction points in a floor plan image. Then the junctions is located for the connection of walls. Only the location of the walls of uniform thickness along XY-pricipal direction are processed \cite{liu2017raster}. A fully convolutional network is proposed to label pixels in a floor plan. However, this method just applies a general segmentation network to recognize pixels of different classes. The spatial relations between the floor plan elements and room boundary are ignored \cite{yamasaki2018apartment}. A context module is developed to extract the features of the spatial relations between the floor plan elements and the room boundary. However, the kernels in the context module is not efficient for the learning of the spatial relations \cite{zeng2019deep}. And the pixel labels of the rooms and other elements are noisy, the performance are beyond satisfactory.

\begin{figure*}
	\centering
	\includegraphics[width=12cm]{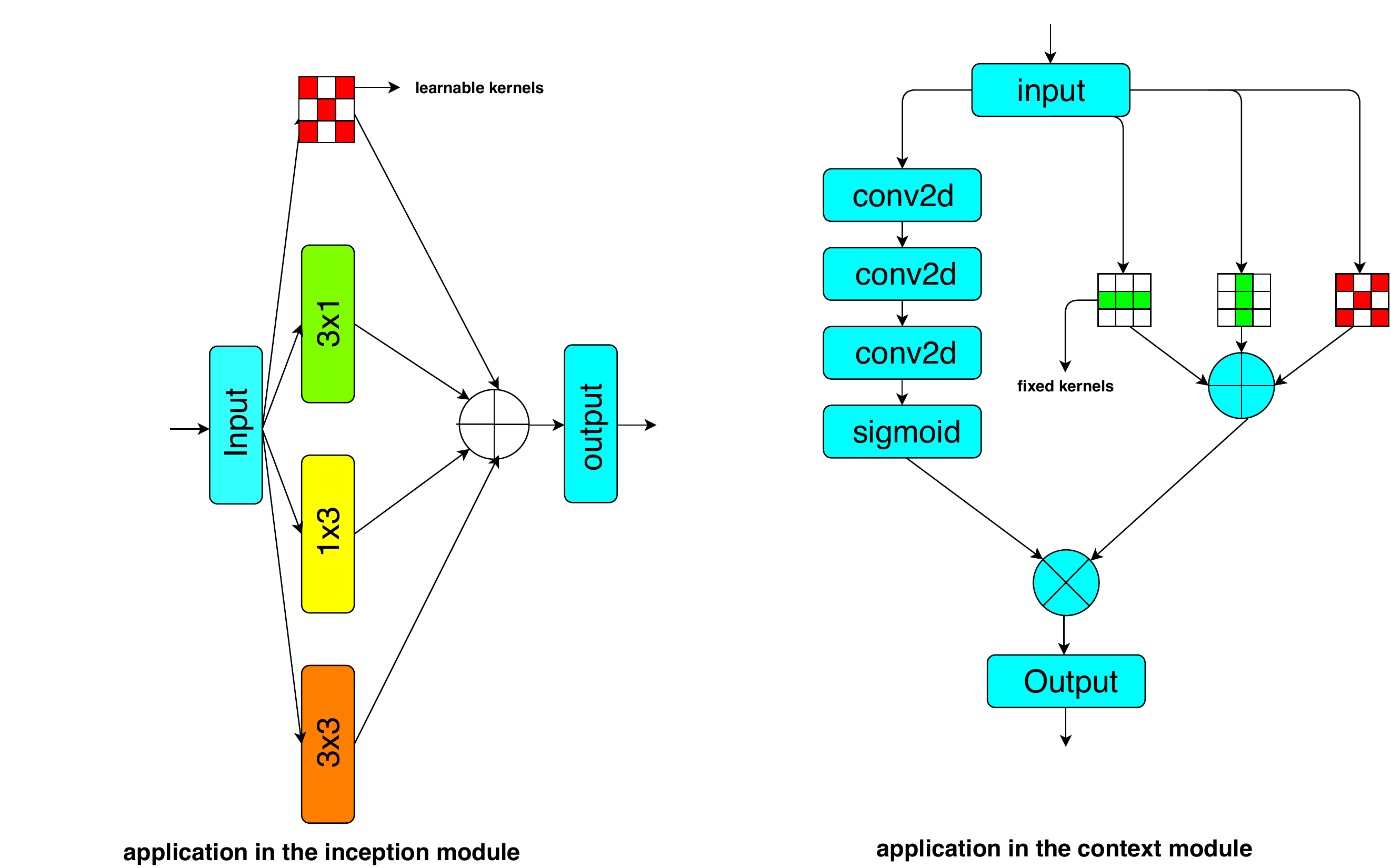}
	\caption{Two application in the inception module and context module.}
	\label{fig3}
\end{figure*}

Therefore, we proposed a new network for the floor plan recognition, focusing on the exploration of spatial relationship among elements with both common and irregular shapes in the floor plan, the reduction of the noise in the pixel labels and the improvement of accuracy of the pixel labels. As Figure \ref{fig1} shows, the recognition of both the boundary, room and other elements with irregular shapes is far beyond satisfactory. 

As these elements are inter-related graphical elements with structural semantics in the floor plans, the backbone network \cite{zeng2019deep} is a multi-task neural network based on the hierarchy of labels. And the direction-aware, learnable, additive \cite{ding2019acnet} kernels are proposed for the sufficient learning of the spatial relations as there is a variety of rooms and basic elements with irregular shapes. As Figure \ref{fig2} shows, the proposed kernels learn to adapt to different direction for elements with the irregular shapes. As Figure \ref{fig3} shows, this proposed kernel is applied in the context module \cite{zeng2019deep} to learn the spatial relations for the irregular shapes. Besides, the proposed kernel is used in the blocks of the backbone network for obtaining higher accuracy of pixel labels. Moreover, the adversarial framework is applied to further improve the accuracy of the pixel labels and reduce the noise of the results, as Figure \ref{fig4} shows.

Our contributions are threefold. First, the proposed kernels are designed with additivity\cite{ding2019acnet}, directional-aware function and learn-able parameters for better understanding of rooms and other elements with both irregular and common shapes in the floor plan. Second, the proposed kernels are applied in both the convolutional blocks and the context modules for the improvement of the accuracy of pixel labels. Thirdly, the adversarial technique \cite{goodfellow2014generative} is applied to both increase the accuracy and reduce the noise of the pixel labels.  

\section{Related Work}
Traditional methods are developed for the recognition of elements in the floor plan. Most of them are based on the low-level image processing. For example, a semi-automatic method is applied for the room segmentation \cite{ryall1995semi}. Walls, doors and rooms are detected by graphical shapes in the layout by other early methods \cite{ah1997variations,dosch2000complete}. Bitmapped floor plans are converted to vector graphics and 3D room models are then generated \cite{or2005highly}. Text is separated and then walls are separated, the door and windows are located finally \cite{ahmed2011improved}. The heuristics are applied to recognize walls and openings, then 3D building models are generated. However, using heuristics and low-level methods to recognize low-level elements in floor plans is error-prone. 
Deep learning methods are used to address this problem \cite{dodge2017parsing, Raster2Vector}. A fully convolutional network(FCN) \cite{dodge2017parsing} is applied firstly to detect the wall pixels, and a faster R-CNN framework \cite{ren2015faster} is applied to detect doors, sliding doors, symbols such as kitchen stoves. Similarly, a deep neural network is firstly identified for the junction points in floor plan image, then integer programming is used to join the junctions to locate the walls in the floor plan. However, these two methods is able to recognize layouts with only rectangular rooms and walls of uniform thickness. Later, a FCN is trained to produce the semantic segmentation of a floor plan \cite{yamasaki2018apartment}, the semantic segmentation is then to form a graph model for the retrieve houses of similar structures. Besides, the spatial relations among classes of different types of rooms is learned through a context module \cite{zeng2019deep}, and a multi-task approach is applied to maximize the learning of the floor plan elements. However, the above deep learning methods are not sufficient for the recognition of the floor plan elements with irregular shapes such as slope walls and circular rooms. And there are noise in the pixel labels of different types of rooms.

The GAN framework is developed \cite{goodfellow2014generative} and its theoretical foundation is built. The methodology of adversarial training has been applied widely, such as image generation \cite{salimans2016improved}, image completion \cite{li2017generative}, super-resolution \cite{ledig2017photo}, object detection \cite{wang2017fast}, domain adaption \cite{hoffman2017cycada} and semantic segmenation \cite{luc2016semantic}. However, to the best of our knowledge, the adversarial technique is not applied in the recognition of floor plan elements yet. And the adversarial network is not yet designed to improve the performance of the recognition task and to reduce the noise of the labeled pixels.   
Recently, several other works \cite{Zhang_2014_ECCV,Lee_2017_ICCV,Zou_2018_CVPR,Yang_2019_CVPR,Sun_2019_CVPR} are exploited and are related to room layouts. Their focus is to reconstruct 3D room layouts from images.

\section{Technical Background}
For a convolutional layer with a kernel size of $ H \times W$ and $D$ filters which takes a $C$ channel feature map as input, let $F \in \mathbb{R}^{H\times W \times C}$ denote the 3D convolution kernel of a filter, let $M \in \mathbb{R}^{U \times V \times C}$ denote the input, $M$ is a feature map with a spatial resolution of $U \times V$ and $C$ channels, and $O \in \mathbb{R}^{U \times V}$ denotes the output of $D$ channels. For the $jth$ filter at such a layer, the corresponding output feature map channel is the following with batch normalizations \cite{ding2019acnet} and linear scaling transformation adopted.

\begin{equation}
    O_{:,:,j} = (\sum_{k=1}^{C}M_{:,:,k} * F_{:,:,k}^{j}-\mu_{j}) * \frac{\gamma_{j}}{\sigma_{j}}+ \beta_{j}
\end{equation}

where $u_{j}$ and $\sigma_{j}$ are the values of channel-wise mean and standard deviation of batch normalization, $\gamma_{j}$ and $\beta_{j}$ are the learned scaling factor and bias term, respectively. Also, $M_{:,:,k}$ is the $kth$ channel of $M$ in the form of a $U \times V$ matrix, and $F^{j}_{:,:,k}$ is the $kth$ input channel of $F^{j}$. 

If the same resolution of the outputs are produced by several 2D kernels with compatible sizes operating on the same input, these kernels on the corresponding positions could be added up to produce the output. And, this additivity may hold for 2D convolutions, even with different kernel sizes.

\begin{figure*}
	\centering
	\includegraphics[width=12cm]{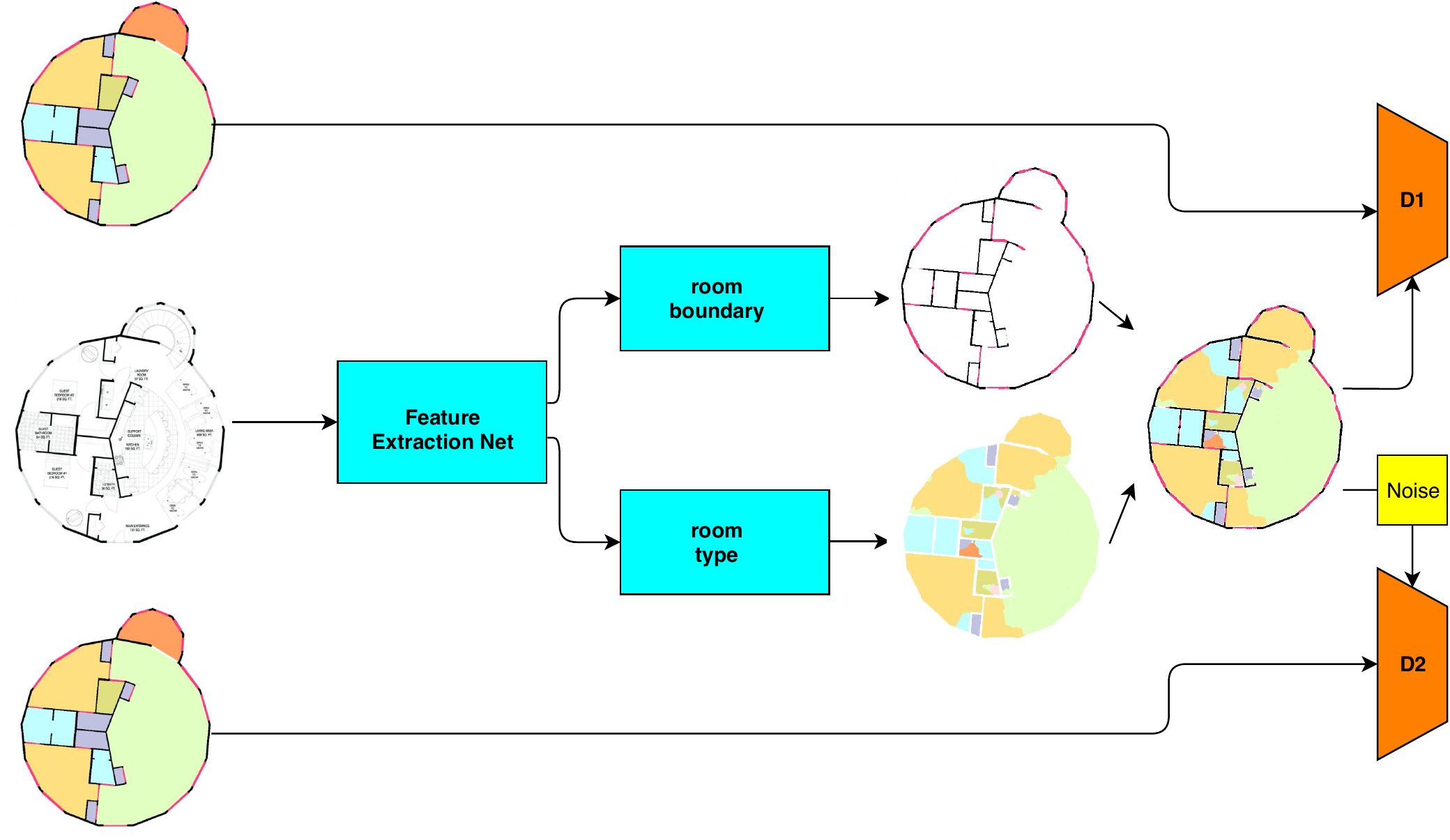}
	\caption{The overall network architecture for the generator and discriminator network.}
	\label{fig4}
\end{figure*}

\begin{equation}
   I \times K^{(1)} + I \times K^{(2)} = I \times (K^{(1)} \bigoplus K^{(2)}) 
\end{equation}
where $I$ is a matrix, $K^{(1)}$ and $K^{(2)}$ are the two 2D kernels with compatible sizes, and $ \bigoplus$ is the element-wise addition of the kernel parameters on the corresponding positions. 

\section{Kernel Formulation}
For the task of recognizing the elements in the floor plan layouts such as walls, doors, windows and different types of rooms, two kernels are approached to recognize both common and irregular walls, rooms and etc al. For example, for the segmentation of circular or inclined walls, rooms with circular or inclined walls, common kernels with the size of $3 \times 3$, $1 \times 3$ or $3 \times 1$ are not efficient enough for the abstraction of features of irregular floor plan elements. Therefore, two kernels $K^{(1)}$ and $K^{(2)}$ are approached as following:

\begin{equation}
K^{(1)} = W_{1},
W_{1}=\left(               
  \begin{array}{ccc}   
    a_{11} & 0 & 0\\  
    0 & a_{22} & 0\\
    0 & 0 & a_{33}\\
  \end{array}
\right)         
\end{equation}

\begin{equation}
K^{(2)} = W_{2},
W_{2}=\left(               
  \begin{array}{ccc}   
    0 & 0 & b_{13}\\  
    0 & b_{22} & 0\\
    b_{31} & 0 & 0\\
  \end{array}
\right)
\end{equation}

where $a_{11}$, $a_{22}$, $a_{33}$, $b_{13}$, $b_{22}$ and $b_{31}$ are six parameters which are learned in the training process, as Figure \ref{fig2} shows. The two kernels are firstly added as one kernel with kernel additivity\cite{ding2019acnet} $K = K^{(1)} + K^{(2)}$, and then the learned parameters is trained to adapt to different direction of elements with both common and irregular shapes.

Where $K = K^{(1)} + K^{(2)}$ is applied in the following two modules for the efficient feature extraction of the floor plan elements. 

\begin{figure*}
	\centering
	\includegraphics[width=12cm]{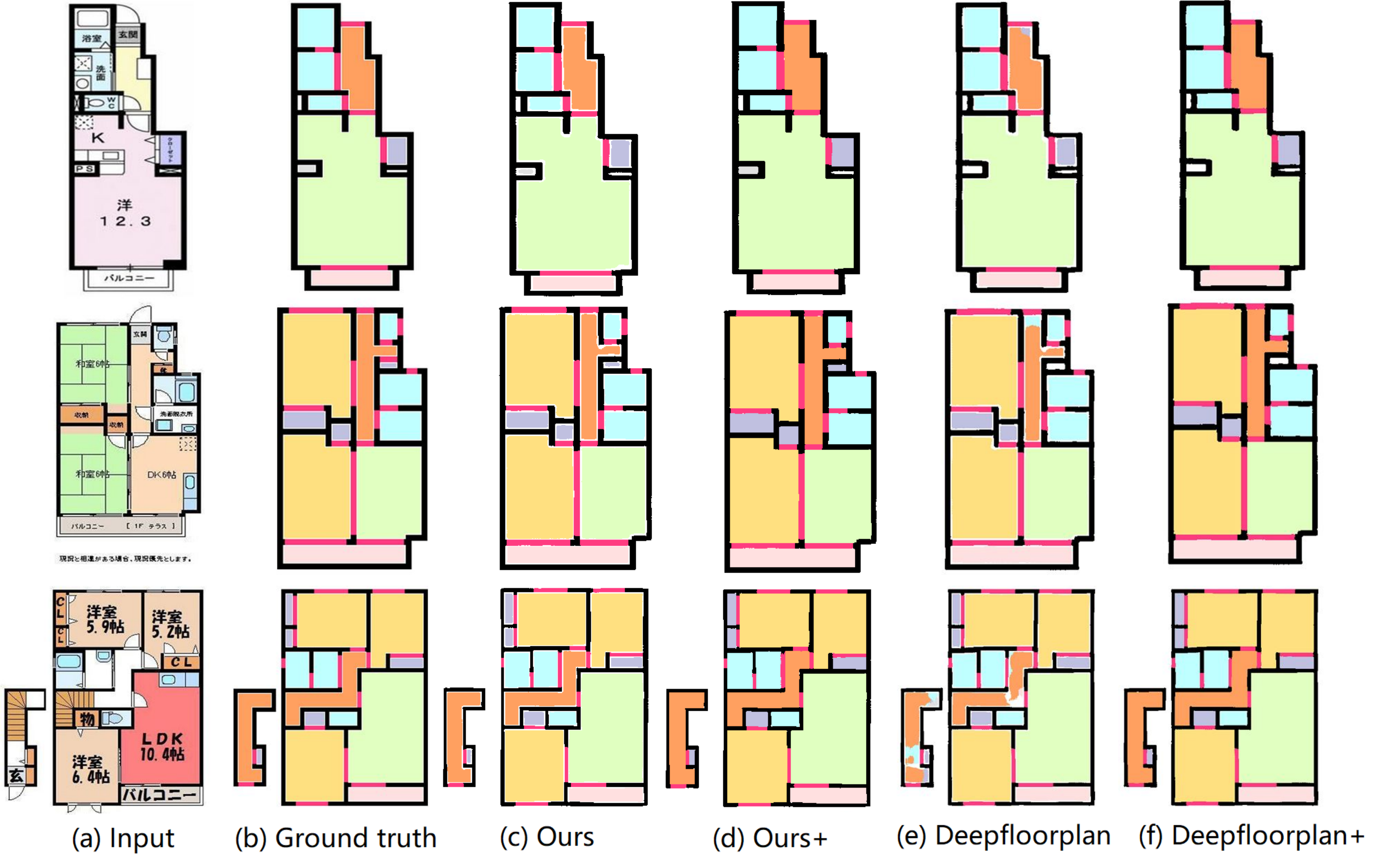}
	\caption{Visual comparison of floor plan recognition between results produced by our methods and others on the R2V. $+$ represents the results after post-processing \cite{zeng2019deep}.}
	\label{fig5}
\end{figure*}

\subsection{Inception Module with Kernels}
The Inception module is widely used in deep neural networks, this module applies different size of kernels. In order to apply this module for the tasks of recognizing both common and irregular walls, rooms and etc al, the kernel $K$ is applied in the module and sum with the output of other size of kernels. As Figure \ref{3} shows, for example, kernels of size $1 \times 3$, $3 \times 1$, $3 \times 3$ and $K$ is added in the inception module. Then, the efficient feature of both common and irregular elements through the proposed kernel $K$ is learned in the backbone network \cite{zeng2019deep}. 

\subsection{Context Module with Kernels}
The context module is applied for the segmentation of rooms with using the context of walls, doors and windows \cite{zeng2019deep}. However, it's not efficient enough for the usage of irregular walls, doors and windows. Therefore, we apply the proposed kernel $K$ in the context module, As Figure \ref{fig3} shows, it's summed with the output of kernels with $1 \times 3$ and $3 \times 1$, 

\section{Adversarial Module}
Two generation networks and two discriminator networks are proposed for the semantic segmentation of the floor plan elements. As Figure \ref{fig4} shows, given an input image with the dimension $H \times W \times 3$, the segmentation network outputs two class probability maps of $H \times W \times C_{1}$ and $H \times W \times C_{2}$, where $H \times W \times C_{1}$ is the segmentation map of the walls, doors and windows, $H \times W \times C_{2}$ is the segmentation map of the rooms. $C_{1}$ is the number of the class of basic floor plan elements and $C_{2}$ is the number of the class of rooms. Two adversarial modules are proposed to improve the performance of recognising the floor-plan elements.

\subsection{Regular Adversarial Module}
As Figure \ref{fig4} shows, the generator in this module is the proposed segmentation network, the discriminator $D_{1}$ is a typical GAN discriminator which takes input images and output a single probability value. During the training process, the segmentation network is supervised by both the cross-entropy loss with the ground truth label map and the adversarial loss with the discriminator network. The adversarial loss is a regular loss.

\begin{figure*}
	\centering
	\includegraphics[width=12cm]{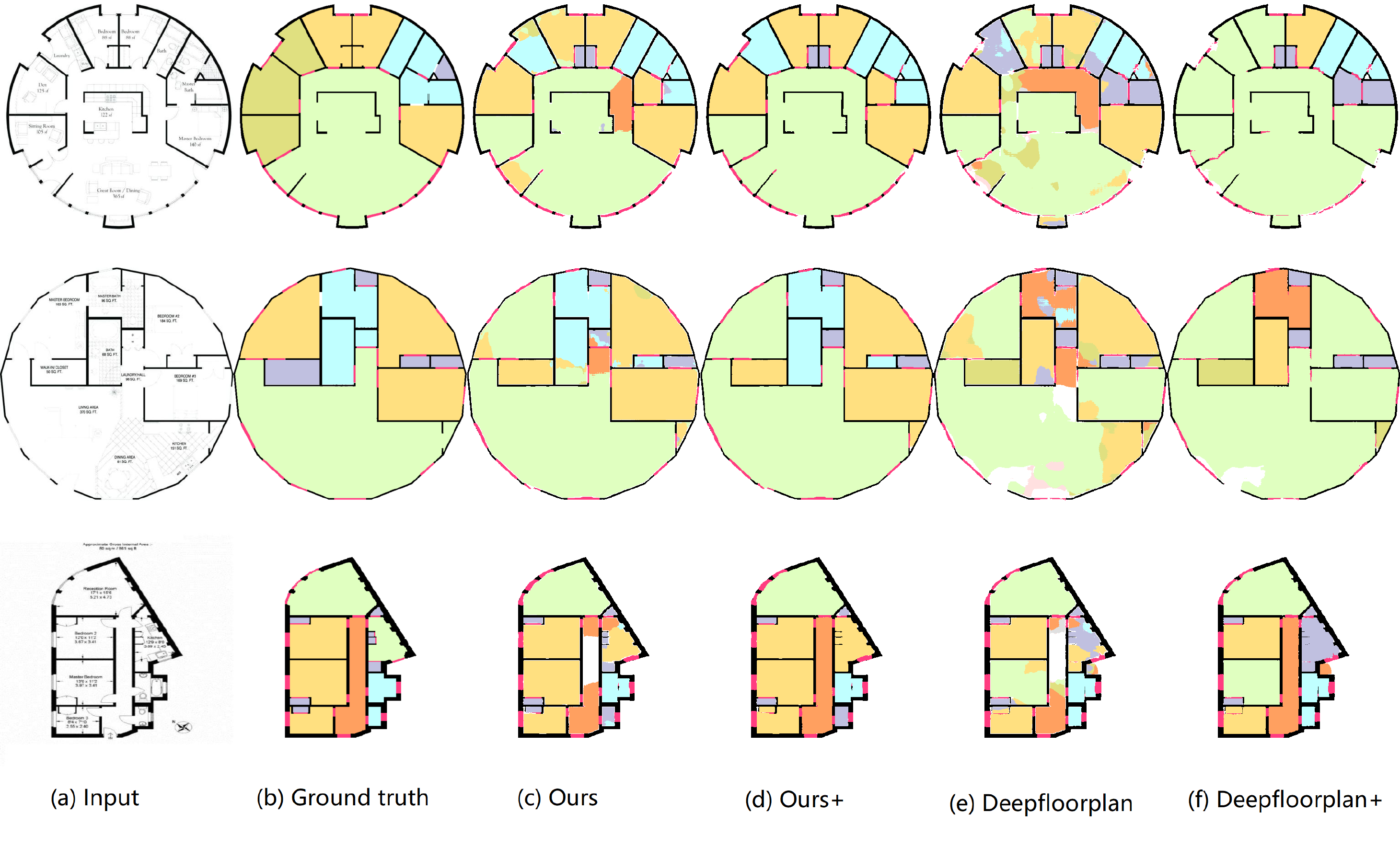}
	\caption{Visual comparison of floor plan recognition between results produced by our methods and others on the R3D. $+$ represents the results after post-processing \cite{zeng2019deep}.}
	\label{fig6}
\end{figure*}

\subsection{Adversarial Noise Module}
As Figure \ref{fig4} shows, the generator in this module produces both the semantic segmentation of floor plan elements, and the noisy map of the semantic segmentation through a noise module. The noise module apply noise with Gaussian random and uniform random distribution for each patch of the segmentation. As shown, it's produced through a yellow noise module. Typical GAN discriminator $D_{2}$ is applied to takes output of the generator and output a single probability value. During the training process, the segmentation network is supervised by both the cross-entropy loss with the ground truth label map and the adversarial loss with the discriminator network. The adversarial loss is a measure of the difference between probability value when the discriminator takes labeled segmentation map and noisy segmentation map from the segmentation network.    

\subsection{Training Objective}
Given an input image $X_{n}$ of size $H \times W \times 3$, the segmentation network is denoted as $S(*)$ and the predicted probability map is denoted as $S_{1}(X_{n})$ of size $H \times W \times C_{1}$ and $S_{2}(X_{n})$ of size $H \times W \times C_{2}$ where $C_{1}$ is the category number of walls, doors and windows, $C_{2}$ is the category number of the rooms. For the first adversarial module, the first adversarial module is denoted as $D_{1}(*)$ which outputs a two-class confidence value. The second adversarial module is denoted as $D_{2}(*)$ which outputs a confidence value for the segmentation map and corresponding noisy map. 

\subsubsection*{Discriminator network training}To train the first discriminator network, the spatial cross-entropy loss $L_{1D}$ is formally written as following:

\begin{figure}
	\centering
	\includegraphics[width=6cm]{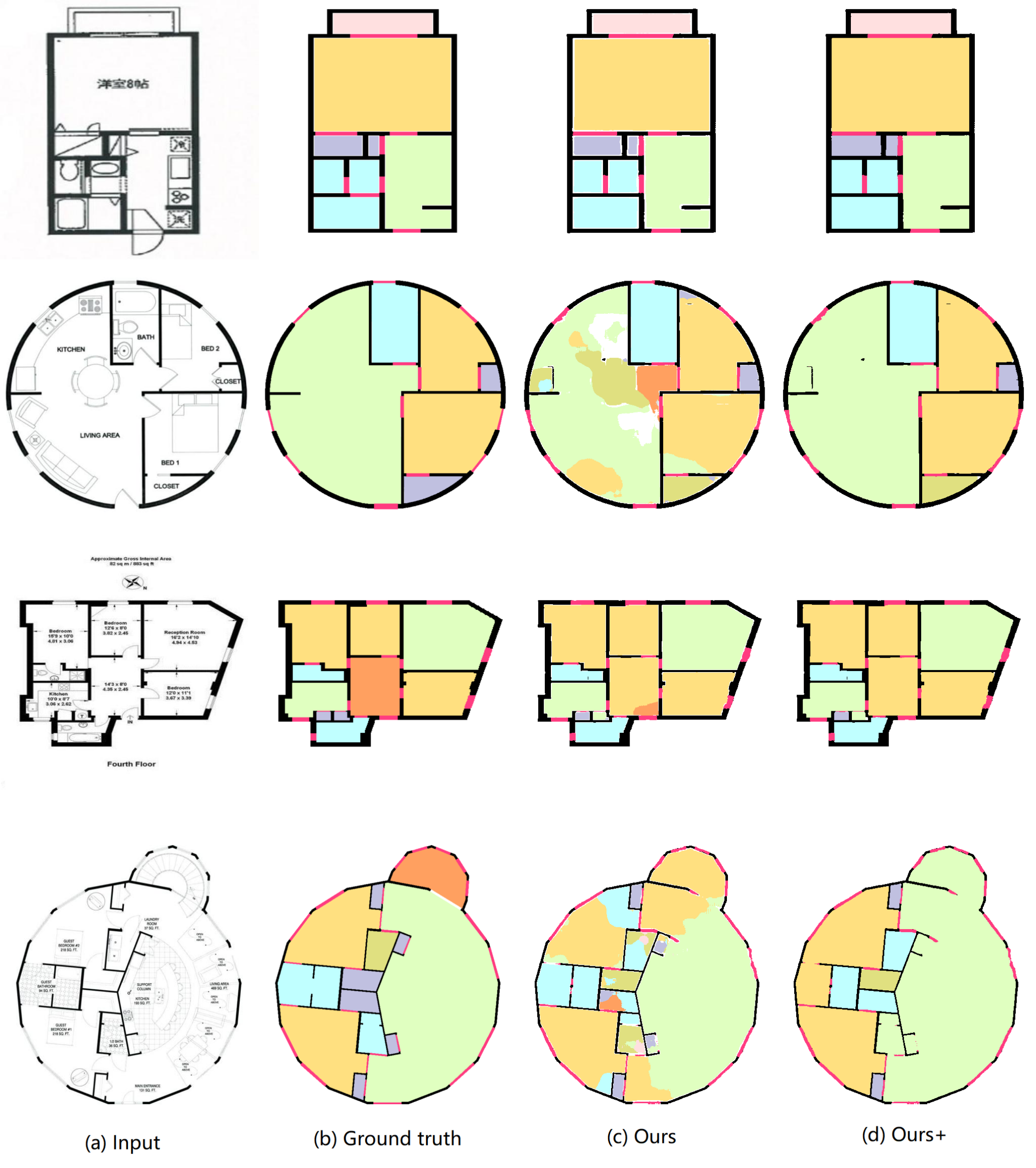}
	\caption{Visual results of irregular shapes of deep floor. $+$ represents the results after post-processing \cite{zeng2019deep}.}
	\label{fig7_1}
\end{figure}

\begin{equation}
L_{1D} = -(1-y_{1n})log(D(P_{n}^{0})) + y_{1n}log(D(P_{n}^{1}))    
\end{equation}

\begin{figure}
	\centering
	\includegraphics[width=6cm]{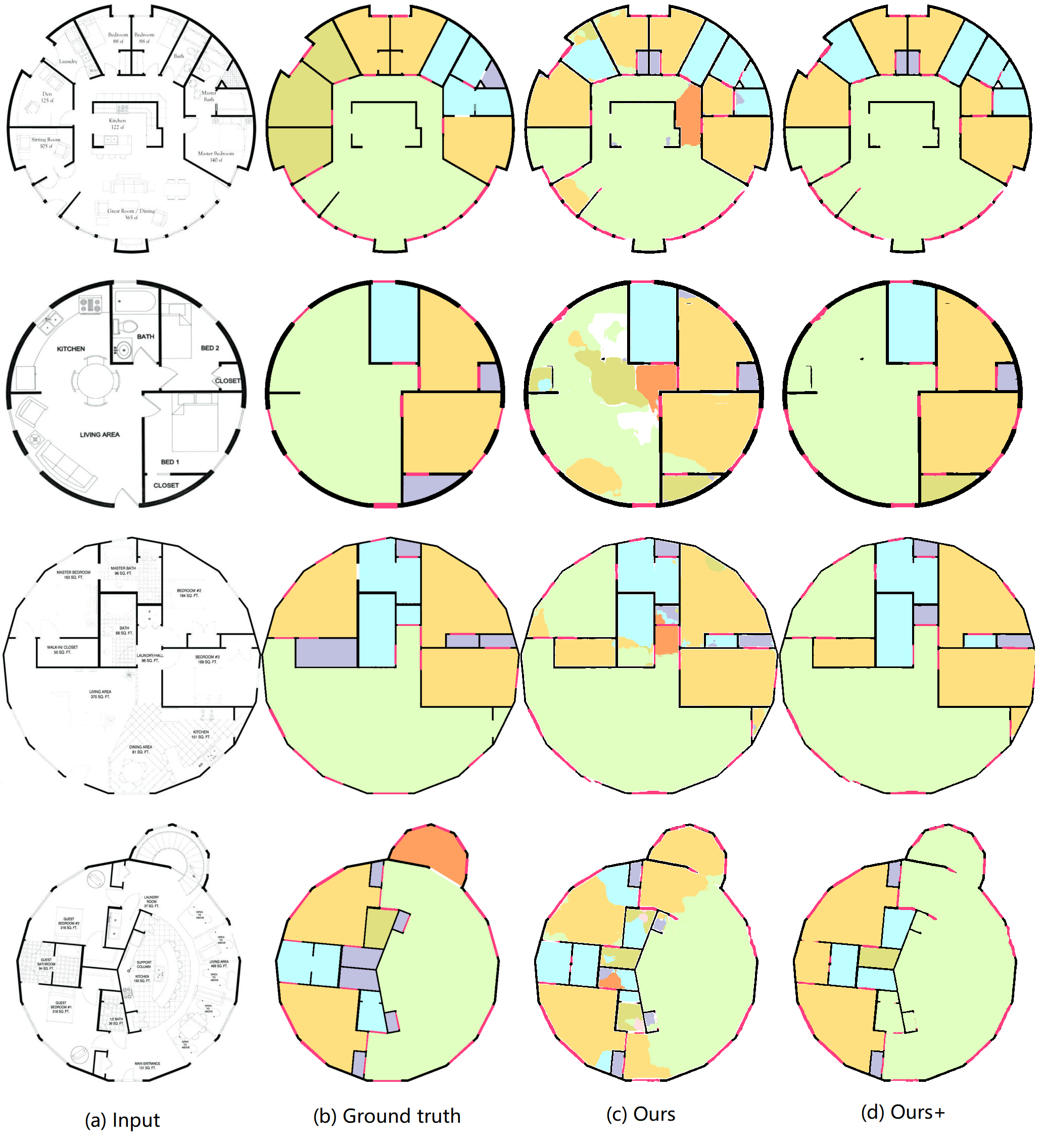}
	\caption{Visual results of irregular shapes of deep floor. $+$ represents the results after post-processing [D].}
	\label{fig7_2}
\end{figure}

where $y_{1n}=0$ if sample $P_{n}^{0}$ is drawn from the segmentation network, and $y_{1n}=1$ if the sample $P_{n}^{1}$ is from the ground truth label. 

To train the second discriminator network, the spatial cross-entropy loss $L_{2D}$ is formally written as following:

\begin{table*}
		\begin{center}
			\begin{tabular}{cccccccccc}
				\hline
				\multicolumn{1}{c}{\multirow{2}{*}{\textbf{Model}}}{\multirow{2}{*}{\textbf{Over-acc}}}&&&&\textbf{Class-acc}\\
                \multicolumn{1}{c}{}{}&Wall&Door-Win&Closet&Bathroom&Livingroom&Bedroom&Hall&Balcony\\
                \hline
                \multicolumn{1}{c}{RVN}{$0.84$}&$0.53$&$0.58$&$0.78$&$0.83$&$0.72$&$0.89$&$0.64$&$0.71$\\
                \hline
                \multicolumn{1}{c}{FP}{$0.88$}&$0.88$&$0.86$&$0.80$&$0.86$&$0.86$&$0.75$&$0.73$&$0.86$\\
                \hline
                \multicolumn{1}{c}{FP+}{$0.89$}&$\textbf{0.88}$&$\textbf{0.86}$&$0.82$&$0.90$&$0.87$&$0.77$&$0.82$&$0.93$\\
                \hline
                \multicolumn{1}{c}{Ours}{$0.91$}&$0.84$&$0.83$&$0.87$&$0.88$&$0.86$&$0.91$&$0.87$&$0.90$\\
                \hline
                \multicolumn{1}{c}{Ours+}{$\textbf{0.92}$}&$0.84$&$0.83$&$\textbf{0.91}$&$\textbf{0.94}$&$\textbf{0.88}$&$\textbf{0.96}$&$\textbf{0.96}$&$\textbf{0.95}$\\
				\hline
			\end{tabular}
			\label{table1}
			\caption{Comparison with the state-of-the-art model Deepfloorplan \cite{zeng2019deep} on the R2V dataset. Symbol + indicates the method with post-processing}
		\end{center}
	\end{table*}
	
\begin{equation}
L_{2D} = - (1-y_{2n})log(D(P_{n-noise}^{0})) + y_{2n}log(D(P_{n}^{1}))      
\end{equation}

where $y_{2n}=0$ if sample $P_{n-noise}^{0}$ is drawn from the segmentation network with noise, and $y_{2n}=1$ if the sample $P_{n}^{1}$ is from the ground truth label.

\subsubsection*{Segmentation network training}To train the segmentation network via minimizing a multi-task loss function:

\begin{equation}
L_{seg} = L_{seg} + \lambda_{adv1}L_{adv1} + \lambda_{adv2}L_{adv2}
\end{equation}

where $L_{seg}$, $L_{adv}$ denote the spatial multi-class entropy loss and the adversarial loss respectively, $\lambda_{adv1}$ and $\lambda_{adv2}$ are two constants for balancing the multi-task training.

Given an input image $X_{n}$, ground truth $Y_{n}$ and prediction results $S_(X_{n})=P_{n}^{0}$, the cross-entropy loss is written as following:

\begin{equation}
L_{seg} = -Y_{n} log(P_{n}^{0})
\end{equation}

And the $L_{adv1}$ and $L_{adv2}$ are written as following:

\begin{equation}
L_{adv1} = L_{adv2} = -log(D(P_{n}^{1}))
\end{equation}

\begin{figure*}
	\centering
	\includegraphics[width=14cm]{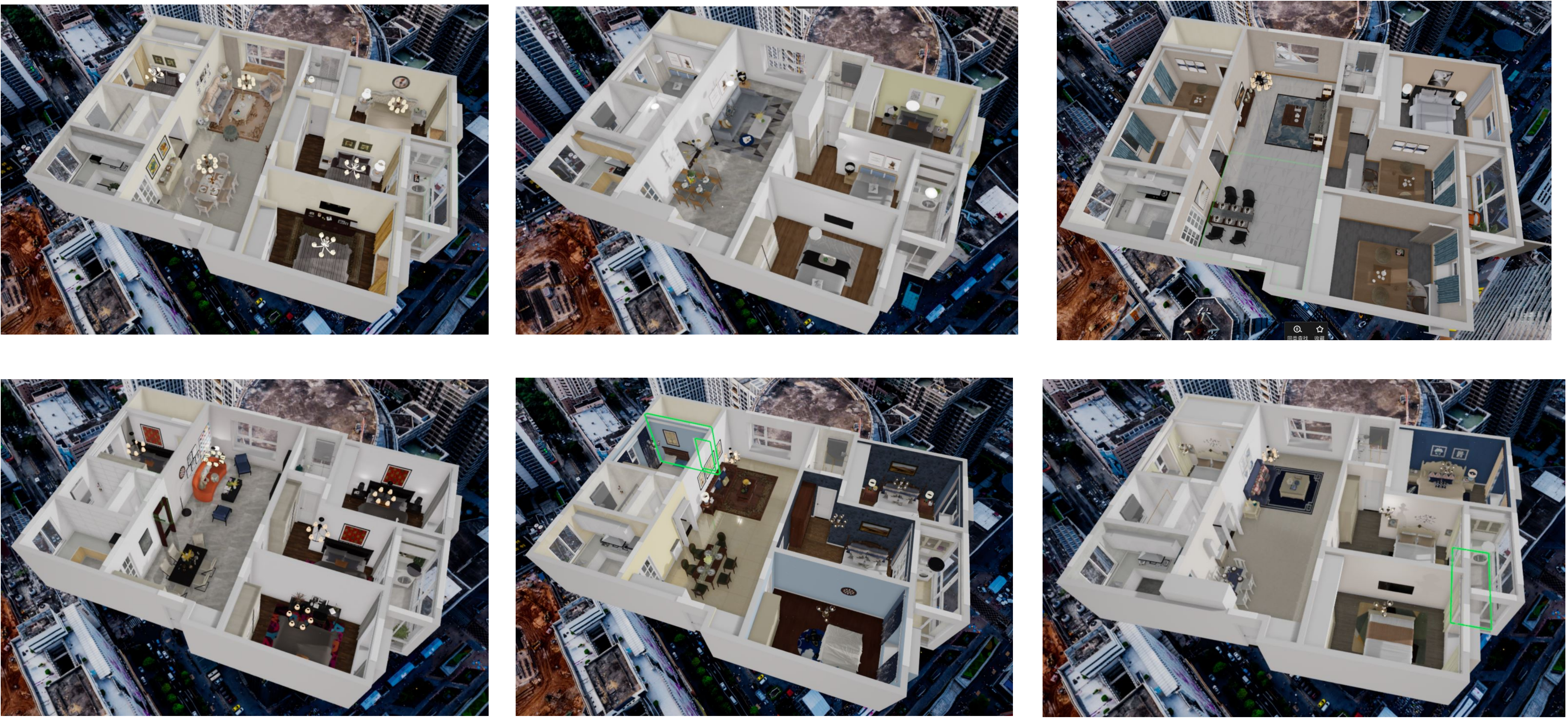}
	\caption{Visual results of the 3D automatic furniture layout from our results. These results are sold at the industrial level. }
	\label{fig8}
\end{figure*}

During the training process, the adversarial loss is used to fool the discriminator by maximizing the probability of the segmentation prediction being considered as the ground truth distribution.

\subsubsection*{Network Architecture}
The segmentation network is the backbone network \cite{zeng2019deep} with applying inception module and context module using the kernel $K$. For the discriminator network, nine residual blocks are applied with one 2D convolution layer in the network. The residual blocks contain $1$, $2$, $3$, $4$, $6$, $6$, $8$, $8$, $8$ 2D convolution layers. Each convolution layer is followed by a Leaky-ReLU parameterized through $0.2$ except the last layer. The last layer is an average pooling layer and is followed by a sigmoid function layer. 

\section{Experiments}
\subsection{Implementation Details}
\subsubsection{Network Training.}
Both the proposed generative network and discriminator network are trained on an NVIDIA TITAN Xp GPU. $60k$ iterations are ran in total. The Adam optimizer is employed to updated the parameters and a fixed learning rate of $1e_{4}$ are applied to train the network. The resolution of the input floor plan is $512 \times 512$, it's used to keep the thin and short lines in the floor plans. For other existing methods in the comparison, the original hyper-parameters of the reported papers are used to train their networks. Also, the results are further evaluated every five training epochs to obtain the best recognition results. 

\subsection{Qualitative and Quantitative Comparisons}
\subsubsection*{Comparing with Deepfloorplan}
First, the proposed network is compared with the deepfloorplan \cite{zeng2019deep}, the state-of-the-art method for the floor plan recognition. Specifically, images from the R2V dataset and R3D are used to train the proposed network and the baseline networks. Considering that the R2V network can only output 2D corner coordinates of bounding boxes, the procedure presented in \cite{liu2017raster} is used to convert its bounding box outputs to per-pixel labels.

Figure \ref{fig5} shows visual comparisons between our proposed network and Deepfloorplan. Both pre-processing and post-processing results are used for the comparison. For the evaluation with the R2V dataset, comparing with the deepfloorplan network \cite{zeng2019deep}, the baseline network tends to have poorer performance on the room-boundary prediction, or miss some room regions. The proposed methods are more similar to the ground truth, even without post-processing. 

For the evaluation with the R3D dataset, it contains many non-rectangular room shapes, so the baseline network produces many missing regions of rooms. The boundary of the circle is hard to recognize. Besides, the rooms with the boundary of the circle is missing or the types of the rooms are wrong. However, results from the proposed network are much better.  

Two widely-used metrics are adopted for the quantitative evaluation on the floor plan tasks. The overall pixel accuracy and the per class pixel accuracy are written as following:

\begin{equation}
class-accuracy(i) = \frac{N_{i}}{N^{'}_{i}}
\end{equation}

\begin{equation}
over-accuracy = \frac{\sum_{i}N_{i}}{\sum_{i}N^{'}_{i}}
\end{equation}

\begin{table*}
		\begin{center}
			\begin{tabular}{ccccccccccc}
				\hline
				\multicolumn{2}{c}{\multirow{2}{*}{\textbf{Model}}}{\multirow{2}{*}{\textbf{Over-acc$\quad$}}}{\multirow{2}{*}{\textbf{M-IOU}}}&&&&\textbf{CA}\\
                \multicolumn{2}{c}{}{}{}{}&Wall&Door-Win&Closet&Bathroom&Living&Bedroom&Hall&Balcony\\
                \hline
                \multicolumn{2}{c}{PSP}{$0.88$/$0.88\quad$}{$0.70$/$0.69$}&$0.84$/$0.84$&$0.76$/$0.76$&$0.80$/$0.71$&$0.90$/$0.84$&$0.83$/$0.90$&$0.86$/$0.92$&$0.78$/$0.81$&$0.87$/$0.82$\\
                \hline
                \multicolumn{2}{c}{DLV3}{$0.88$/$0.87\quad$}{$0.69$/$0.67$}&$0.80$/$0.80$&$0.72$/$0.72$&$0.78$/$0.85$&$0.90$/$0.90$&$0.85$/$0.84$&$0.82$/$0.65$&$0.55$/$0.87$&$0.87$/$0.45$\\
                \hline
                \multicolumn{2}{c}{FP}{$0.89$/$0.90$\quad}{$0.74$/$0.76$}&$0.89$/$0.89$&$0.89$/$0.89$&$0.81$/$0.92$&$0.87$/$0.93$&$0.88$/$0.91$&$0.83$/$0.91$&$0.68$/$0.84$&$0.90$/$0.92$\\
                \hline
                \multicolumn{2}{c}{Ours}{$0.91$/$\textbf{0.92}$\quad}{$\textbf{0.81}$/$0.81$}&$0.84$/$0.84$&$0.83$/$0.83$&$0.84$/$\textbf{0.91}$&$0.88$/$\textbf{0.94}$&$0.86$/$0.88$&$0.91$/$\textbf{0.96}$&$0.87$/$\textbf{0.96}$&$0.90$/$\textbf{0.95}$\\
                \hline
			\end{tabular}
			\caption{Comparison with PSPnet\cite{zhao2017pyramid},DeeplabV3\cite{chen2018encoder} Deepfloorplan \cite{zeng2019deep} and ours on the R2V dataset. for both pre-processing and post-processing. To be noted, CA denotes class accuracy, M-IOU denotes mean IOU, Over-acc denotes overall accuracy.}
		\end{center}
		\label{table2}
	\end{table*}

\begin{table*}
		\begin{center}
			\begin{tabular}{ccccccccccc}
				\hline
				\multicolumn{2}{c}{\multirow{2}{*}{\textbf{Model}}}{\multirow{2}{*}{\textbf{Over-acc$\quad$}}}{\multirow{2}{*}{\textbf{M-IOU}}}&&&&\textbf{CA}\\
                \multicolumn{2}{c}{}{}{}{}&Wall&Door-Win&Closet&Bathroom&Living&Bedroom&Hall&Balcony\\
                \hline
                \multicolumn{2}{c}{PSP}{$0.84$/$0.81\quad$}{$0.50$/$0.41$}&$0.91$/$0.91$&$0.54$/$0.54$&$0.45$/$0.09$&$0.70$/$0.50$&$0.76$/$0.89$&$0.55$/$0.40$&$0.61$/$0.23$&$0.41$/$0.11$\\
                \hline
                \multicolumn{2}{c}{DLV3}{$0.85$/$0.83\quad$}{$0.50$/$0.44$}&$0.93$/$0.93$&$0.60$/$0.60$&$0.24$/$0.05$&$0.76$/$0.57$&$0.76$/$0.90$&$0.56$/$0.40$&$0.72$/$0.44$&$0.08$/$0.00$\\
                \hline
                \multicolumn{2}{c}{FP}{$0.89$/$0.90$\quad}{$0.63$/$0.66$}&$0.98$/$0.98$&$0.83$/$0.83$&$0.61$/$0.54$&$0.81$/$0.78$&$0.87$/$0.93$&$0.75$/$0.79$&$0.59$/$0.68$&$0.44$/$0.49$\\
                \hline
                \multicolumn{2}{c}{Ours}{$0.94$/$\textbf{0.94}$\quad}{$0.74$/$\textbf{0.75}$}&$0.95$/$\textbf{0.95}$&$0.82$/$\textbf{0.82}$&$\textbf{0.65}$/$0.57$&$\textbf{0.93}$/$0.87$&$0.91$/$\textbf{0.98}$&$0.86$/$\textbf{0.86}$&$0.80$/$\textbf{0.87}$&$\textbf{0.76}$/$0.69$\\
                \hline
			\end{tabular}
			\caption{Comparison with PSPnet\cite{zhao2017pyramid},DeeplabV3\cite{chen2018encoder} Deepfloorplan\cite{zeng2019deep} and ours on the R3D dataset, for both pre-processing and post-processing.}
		\end{center}
		\label{table3}
	\end{table*}
	
where $N^{'}_{i}$ and $N_{i}$ are the total number of the ground-truth pixels and the correctly-predicted pixels for the $ith$ floor plan element, respectively. Table \ref{table1} shows the quantitative comparison results on the R2V dataset. From the evaluation results, the proposed method achieves higher accuracy for the floor plan elements, and the post-processing results further improve the performance. In detail, as Figure \ref{fig5} shows, the proposed network shows better results than the baseline network \cite{zeng2019deep}.  

\subsubsection*{Comparing with Segmentation Networks.}
We also make comparison with the state-of-the-art segmentation networks such as DeepLabV3+ \cite{chen2018encoder} and PSPNet\cite{zhao2017pyramid}. The baseline networks are trained on the R2V dataset and the R3D dataset, the parameters are adjusted to obtain the best recognition results. The quantitative comparison results for various methods with and without post-processing are reported in the Table \ref{table2} and Table \ref{table3}, for the measure of the class accuracy and overall accuracy. The datasets for the evaluation are the R2V and R3D datasets, in the comparison with deeplabV3+, pspnet and the state-of-the-art deepfloorplan \cite{zeng2019deep},  the proposed network produces floor plan elements with better quality. Particularly, the boundary of the circular rooms or irregular rooms is close to the ground truth Figure \ref{fig6}, Figure \ref{fig7_1} and Figure \ref{fig7_2}. Besides, the class accuracy of the rooms with regular shapes is higher than the baseline networks. Even without post-processing, the proposed network  produces higher accuracy than the baseline networks. 

\section{Discussion.}
\subsection{Application: Automatic Architecturalization of the Interior Design.}
The results of the floor plan recognition is taken to reconstruct 3D models and 3D automatic furniture layouts. As Figure \ref{fig8} shows that several 3D floor plans are constructed. The proposed method is able to recognize walls, rooms and other elements with irregular shapes. Then, 3D walls and different rooms are constructed. Furthermore, the automatic layout of furniture for the interior decoration are based on the recognition \ref{fig8}. As Figure \ref{fig8} shows, the 3D automatic layout are shown. There are only six samples of the hundreds of thousands industrial products.

\subsection{Limitation.}
Here, several challenging situations are discussed. First, the proposed network fails to produce good results of villa when there are several images of different floors in one figure. Second, the proposed method can not produce good semantic segmentation of both the basic elements such as walls, windows, rooms etc al and Icons such as sink, fire place, bathtub etc al. As these icons are important icons for the automatic layout of the furniture, lamps, decorations etc al. 

\section{Conclusion.}
Proposed kernels and adversarial training process with two discriminators are applied in the task of the floor plan. There are three key contributions in this work. First, the direction-aware, learnable, additive kernels are proposed and applied in the context module and the convolutional blocks. These kernels are evaluated to improve the performance of elements in both common and irregular shapes. Second, the adversarial training technique is taken and two discriminators are applied to further improve the accuracy and reduce the noise in the pixel labels. Third, the proposed network is applied in the industrial products as the principal step for the automatic layout of the interior decorations. Up to now, it's used for selling above 100 thousands 3D interior decorations solutions to the proprietors.          

{\small
\bibliographystyle{ieee_fullname}
\bibliography{egbib}

\begin{thebibliography}{10}\itemsep=-1pt

\bibitem{ah1997variations}
Christian Ah-Soon and Karl Tombre.
\newblock {V}ariations on the analysis of architectural drawings.
\newblock In {\em International Conference on Document Analysis and Recognition
  (ICDAR)}. IEEE, 1997.

\bibitem{ahmed2011improved}
Sheraz Ahmed, Marcus Liwicki, Markus Weber, and Andreas Dengel.
\newblock Improved automatic analysis of architectural floor plans.
\newblock In {\em 2011 International Conference on Document Analysis and
  Recognition}, pages 864--869. IEEE, 2011.

\bibitem{chen2018encoder}
Liang-Chieh Chen, Yukun Zhu, George Papandreou, Florian Schroff, and Hartwig
  Adam.
\newblock Encoder-decoder with atrous separable convolution for semantic image
  segmentation.
\newblock In {\em Proceedings of the European conference on computer vision
  (ECCV)}, pages 801--818, 2018.

\bibitem{ding2019acnet}
Xiaohan Ding, Yuchen Guo, Guiguang Ding, and Jungong Han.
\newblock Acnet: Strengthening the kernel skeletons for powerful cnn via
  asymmetric convolution blocks.
\newblock In {\em Proceedings of the IEEE International Conference on Computer
  Vision}, pages 1911--1920, 2019.

\bibitem{dodge2017parsing}
Samuel Dodge, Jiu Xu, and Bj{\"o}rn Stenger.
\newblock {P}arsing floor plan images.
\newblock In {\em International Conference on Machine Vision Applications
  (MVA)}. IEEE, 2017.

\bibitem{dosch2000complete}
Philippe Dosch, Karl Tombre, Christian Ah-Soon, and G{\'e}rald Masini.
\newblock {A} complete system for the analysis of architectural drawings.
\newblock {\em International Journal on Document Analysis and Recognition},
  3(2):102--116, 2000.

\bibitem{gimenez2016automatic}
Lucile Gimenez, Sylvain Robert, Fr{\'e}d{\'e}ric Suard, and Khaldoun Zreik.
\newblock Automatic reconstruction of 3d building models from scanned 2d floor
  plans.
\newblock {\em Automation in Construction}, 63:48--56, 2016.

\bibitem{goodfellow2014generative}
Ian Goodfellow, Jean Pouget-Abadie, Mehdi Mirza, Bing Xu, David Warde-Farley,
  Sherjil Ozair, Aaron Courville, and Yoshua Bengio.
\newblock Generative adversarial nets.
\newblock In {\em Advances in neural information processing systems}, pages
  2672--2680, 2014.

\bibitem{hoffman2017cycada}
Judy Hoffman, Eric Tzeng, Taesung Park, Jun-Yan Zhu, Phillip Isola, Kate
  Saenko, Alexei~A Efros, and Trevor Darrell.
\newblock Cycada: Cycle-consistent adversarial domain adaptation.
\newblock {\em arXiv preprint arXiv:1711.03213}, 2017.

\bibitem{ledig2017photo}
Christian Ledig, Lucas Theis, Ferenc Husz{\'a}r, Jose Caballero, Andrew
  Cunningham, Alejandro Acosta, Andrew Aitken, Alykhan Tejani, Johannes Totz,
  Zehan Wang, et~al.
\newblock Photo-realistic single image super-resolution using a generative
  adversarial network.
\newblock In {\em Proceedings of the IEEE conference on computer vision and
  pattern recognition}, pages 4681--4690, 2017.

\bibitem{Lee_2017_ICCV}
Chen-Yu Lee, Vijay Badrinarayanan, Tomasz Malisiewicz, and Andrew Rabinovich.
\newblock {R}oom{N}et: {E}nd-to-end room layout estimation.
\newblock In {\em IEEE International Conference on Computer Vision (ICCV)},
  2017.

\bibitem{li2017generative}
Yijun Li, Sifei Liu, Jimei Yang, and Ming-Hsuan Yang.
\newblock Generative face completion.
\newblock In {\em Proceedings of the IEEE Conference on Computer Vision and
  Pattern Recognition}, pages 3911--3919, 2017.

\bibitem{liu2017raster}
Chen Liu, Jiajun Wu, Pushmeet Kohli, and Yasutaka Furukawa.
\newblock Raster-to-vector: revisiting floorplan transformation.
\newblock In {\em Proceedings of the IEEE International Conference on Computer
  Vision}, pages 2195--2203, 2017.

\bibitem{Raster2Vector}
Chen Liu, Jiajun Wu, Pushmeet Kohli, and Yasutaka Furukawa.
\newblock {R}aster-to-{V}ector: {R}evisiting floorplan transformation.
\newblock In {\em IEEE International Conference on Computer Vision (ICCV)},
  2017.

\bibitem{luc2016semantic}
Pauline Luc, Camille Couprie, Soumith Chintala, and Jakob Verbeek.
\newblock Semantic segmentation using adversarial networks.
\newblock {\em arXiv preprint arXiv:1611.08408}, 2016.

\bibitem{mace2010system}
S{\'e}bastien Mac{\'e}, Herv{\'e} Locteau, Ernest Valveny, and Salvatore
  Tabbone.
\newblock A system to detect rooms in architectural floor plan images.
\newblock In {\em Proceedings of the 9th IAPR International Workshop on
  Document Analysis Systems}, pages 167--174. ACM, 2010.

\bibitem{or2005highly}
Siu-Hang Or, Kin-Hong Wong, Ying-Kin Yu, and Michael Ming-Yuan Chang.
\newblock {H}ighly automatic approach to architectural floorplan image
  understanding and model generation.
\newblock In {\em Proc. of Vision, Modeling, and Visualization 2005
  (VMV-2005)}, pages 25--32, 2005.

\bibitem{ren2015faster}
Shaoqing Ren, Kaiming He, Ross Girshick, and Jian Sun.
\newblock Faster r-cnn: Towards real-time object detection with region proposal
  networks.
\newblock In {\em Advances in neural information processing systems}, pages
  91--99, 2015.

\bibitem{ryall1995semi}
Kathy Ryall, Stuart Shieber, Joe Marks, and Murray Mazer.
\newblock {S}emi-automatic delineation of regions in floor plans.
\newblock In {\em International Conference on Document Analysis and Recognition
  (ICDAR)}. IEEE, 1995.

\bibitem{salimans2016improved}
Tim Salimans, Ian Goodfellow, Wojciech Zaremba, Vicki Cheung, Alec Radford, and
  Xi Chen.
\newblock Improved techniques for training gans.
\newblock In {\em Advances in neural information processing systems}, pages
  2234--2242, 2016.

\bibitem{Sun_2019_CVPR}
Cheng Sun, Chi-Wei Hsiao, Min Sun, and Hwann-Tzong Chen.
\newblock {H}orizon{N}et: Learning room layout with 1{D} representation and
  pano stretch data augmentation.
\newblock In {\em IEEE Conference on Computer Vision and Pattern Recognition
  (CVPR)}, 2019.

\bibitem{wang2017fast}
Xiaolong Wang, Abhinav Shrivastava, and Abhinav Gupta.
\newblock A-fast-rcnn: Hard positive generation via adversary for object
  detection.
\newblock In {\em Proceedings of the IEEE Conference on Computer Vision and
  Pattern Recognition}, pages 2606--2615, 2017.

\bibitem{yamasaki2018apartment}
Toshihiko Yamasaki, Jin Zhang, and Yuki Takada.
\newblock Apartment structure estimation using fully convolutional networks and
  graph model.
\newblock In {\em Proceedings of the 2018 ACM Workshop on Multimedia for Real
  Estate Tech}, pages 1--6. ACM, 2018.

\bibitem{Yang_2019_CVPR}
Shang-Ta Yang, Fu-En Wang, Chi-Han Peng, Peter Wonka, Min Sun, and Hung-Kuo
  Chu.
\newblock {D}u{L}a-{N}et: A dual-projection network for estimating room layouts
  from a single {RGB} panorama.
\newblock In {\em IEEE Conference on Computer Vision and Pattern Recognition
  (CVPR)}, 2019.

\bibitem{zeng2019deep}
Zhiliang Zeng, Xianzhi Li, Ying~Kin Yu, and Chi-Wing Fu.
\newblock Deep floor plan recognition using a multi-task network with
  room-boundary-guided attention.
\newblock In {\em Proceedings of the IEEE International Conference on Computer
  Vision}, pages 9096--9104, 2019.

\bibitem{Zhang_2014_ECCV}
Yinda Zhang, Shuran Song, Ping Tan, and Jianxiong Xiao.
\newblock {P}ano{C}ontext: {A} whole-room 3{D} context model for panoramic
  scene understanding.
\newblock In {\em European Conference on Computer Vision (ECCV)}, 2014.

\bibitem{zhao2017pyramid}
Hengshuang Zhao, Jianping Shi, Xiaojuan Qi, Xiaogang Wang, and Jiaya Jia.
\newblock Pyramid scene parsing network.
\newblock In {\em Proceedings of the IEEE conference on computer vision and
  pattern recognition}, pages 2881--2890, 2017.

\bibitem{Zou_2018_CVPR}
Chuhang Zou, Alex Colburn, Qi Shan, and Derek Hoiem.
\newblock {L}ayout{N}et: {R}econstructing the 3{D} room layout from a single
  {RGB} image.
\newblock In {\em IEEE Conference on Computer Vision and Pattern Recognition
  (CVPR)}, 2018.

\end{thebibliography}
}

\end{document}